\documentclass[11pt,a4paper]{article}
\usepackage[hyperref]{acl2019}
\usepackage{times}
\usepackage{latexsym}
\usepackage{todonotes}
\usepackage{url}
\usepackage{amsmath}
\usepackage{amssymb}
\usepackage{amsthm}
\aclfinalcopy 
\def\aclpaperid{1399} 

\title{Pretraining Methods for Dialog Context Representation Learning}

\author{Shikib Mehri\thanks{*~Equal contribution.}~, Evgeniia Razumovskaia$^*$, Tiancheng Zhao and Maxine Eskenazi \\
  Language Technologies Institute, Carnegie Mellon University \\
  {\tt \{amehri,erazumov,tianchez,max+\}@cs.cmu.edu}} 

\date{}

\begin{document}
\maketitle
\begin{abstract}
This paper examines various unsupervised pretraining objectives for learning dialog context representations. Two novel methods of pretraining dialog context encoders are proposed, and a total of four methods are examined. Each pretraining objective is fine-tuned and evaluated on a set of downstream dialog tasks using the MultiWoz dataset and strong performance improvement is observed. Further evaluation shows that our pretraining objectives result in not only better performance, but also better convergence, models that are less data hungry and have better domain \linebreak generalizability.
\end{abstract}

\section{Introduction}
Learning meaningful representations of multi-turn dialog contexts is the cornerstone of dialog systems. In order to generate an appropriate response, a system must be able to aggregate information over multiple turns, such as estimating a belief state over user goals~\cite{williams2013dialog} and resolving anaphora co--references~\cite{mitkov2014anaphora}. In the past, significant effort has gone into developing better neural dialog architectures to improve context modeling given the same in-domain training data~\cite{dhingra2016gated,zhou2016multi}. Recent advances in pretraining on massive amounts of text data have led to state-of-the-art results on a range of natural language processing (NLP) tasks~\cite{peters2018deep,radford2018improving,devlin2018bert} including natural language inference, question answering and text classification. These promising results suggest a new direction for improving context modeling by \linebreak creating general purpose natural language representations that are useful for many different downstream tasks.

Yet pretraining methods are still in their infancy. We do not yet fully understand their properties. For example, many pretraining methods are variants of language modeling~\cite{howard2018universal,radford2018improving,devlin2018bert}, e.g. predicting the previous word, next word or the masked word, given the sentence context. This \linebreak approach treats natural language as a simple stream of word tokens. It relies on a complex model to discover high-level dependencies, through the use of massive corpora and expensive computation. Recently the BERT model~\cite{devlin2018bert} achieved state-of-the-art performance on several NLP benchmarks. It introduces a sentence-pair level pretraining objective, i.e. predicting whether two sentences should come after one another. This is a step towards having pretraining objectives that explicitly consider and leverage discourse-level relationships. However, it is still unclear whether language modeling is the most effective method of pretrained language representation, especially for tasks that need to exploit multi-turn dependencies, e.g. dialog context modeling. \citet{thornbury2006conversation} underline several discourse-level features which distinguish dialog from other types of text. Dialog must be coherent across utterance and a sequence of turns should achieve a communicative purpose. Further, dialog is interactive in nature, with feedback and back-channelling between speakers, and turn-taking. 
These unique features of dialog suggest that modelling dialog contexts requires pretraining methods specifically designed for dialog.

Building on this prior research, the goal of this paper is to study various methods of pretraining discourse-level language representations, i.e. modeling the relationship amongst multiple utterances. This paper takes a first step in the creation of a systematic analysis framework of pretraining methods for dialog systems. Concretely, we pretrain a hierarchical dialog encoder~\cite{serban2016building} with four different unsupervised pretraining objectives. Two of the objectives, next-utterance generation~\cite{vinyals2015neural} and retrieval~\cite{lowe2016evaluation}, have been explored in previous work. The other two pretraining objectives, masked-utterance retrieval and inconsistency identification, are novel. The pretrained \linebreak  dialog encoder is then evaluated on several downstream tasks that probe the quality of the learned context representation by following the typical pretrain \& fine-tune procedure.  


Pretraining and downstream evaluation use the MultiWoz dialog dataset~\cite{budzianowski2018multiwoz}, which contains over 10,000 dialogs spanning 6 different domains. The downstream tasks include next-utterance generation (NUG), next-utterance retrieval (NUR), dialog act prediction (DAP), and belief state prediction (BSP). The pretraining objectives are assessed under four different hypotheses: (1) that pretraining will improve downstream tasks with fine-tuning on the entire available data, (2) that pretraining will result in better convergence, (3) that pretraining will perform strongly with limited data and (4) that pretraining facilitates domain generalizability. The results here show that pretraining achieves significant performance gains with respect to these hypotheses. Furthermore, the novel objectives achieve performance that is on-par with or better than the pre-existing methods. The contributions of this paper are: (1) a study of four different pretraining objectives for dialog context representation, including two novel objectives. (2) a comprehensive analysis of the effects of pretraining on dialog context representations, assessed on four different downstream tasks.

\section{Related Work}
This work is closely related to research in auxi-liary multi-task learning and transfer learning with pretraining for NLP systems.

\subsection*{Training with Auxiliary Tasks}
Incorporating a useful auxiliary loss function to complement the primary objective has been shown to improve the performance of deep neural network models, including, but not limited to, error detection \citep{rei2017errordetection}, cross-lingual speech tagging \citep{plank2016multilingual}, domain independent sentiment classification \citep{yu2016sentimentclassification}, latent variable inference for dialog generation~\cite{zhao2017learning} and opinion extraction \citep{ding2017opinionextraction}. Some auxiliary loss functions are designed to improve performance on a specific task. For instance, \citet{yu2016sentimentclassification} pretrained a model for sentiment classification with the auxiliary task of identifying whether a negative or positive word occurred in the sentence. In some cases, auxiliary loss is created to encourage a model's general representational power. \citet{trinh2018longtermdependecies} found that a model can capture far longer dependencies when pretrained with a suitable auxiliary task. This paper falls in line with the second goal by creating learning objectives that improve a representation to capture general-purpose information.
\vspace{-0.3em}
\subsection*{Transfer Learning with Pretraining}

The second line of related research concerns the creation of transferable language representation via pretraining. The basic procedure is typically to first pretrain a powerful neural encoder on massive text data with unsupervised objectives. The second step is to fine-tune this pretrained model on a specific downstream task using a much smaller in-domain dataset~\cite{howard2018universal}. Recently, several papers that use this approach have achieved significant results. ELMo~\cite{peters2018deep} trained a two-way language model with Bidirectional Long Short-Term Memory Networks (biLSTM)~\cite{huang2015bidirectional} to predict both the next and previous word. OpenAI's GPT created a unidirectional language model \linebreak using transformer networks~\cite{radford2018improving} and BERT was trained with two simultaneous objectives: the masked language model and next sentence prediction~\cite{devlin2018bert}. Each of the models has demonstrated state-of-the-art results on the GLUE benchmark \citep{wang2018glue}. The GPT model has also been adapted to improve the performance of end-to-end dialog models. In the 2nd ConvAI challenge \citep{dinan2019second}, the best models on both human and automated eval- uations were generative transformers \citep{wolf2019transfertransfo}, which were initialized with the weights of the GPT model and fine-tuned on in-domain dialog data. These models, which leveraged large-scale pretraining, outperformed the systems which only used in-domain data. 

There has been little work on pretraining methods that learn to extract discourse level information from the input text. Next sentence prediction loss in BERT \citep{devlin2018bert} is a step in this direction. While these pretraining methods excel at modelling sequential text, they do not explicitly consider the unique discourse-level features of dialog. We therefore take the first steps in the study of pretraining objectives that extract better discourse-level representations of dialog contexts. 

\section{Pretraining Objectives}
This section discusses the unsupervised pretraining objectives, including two novel approaches aimed at capturing better representations of dialog context. When considering a specific pretraining method, both the pretraining objective and the model architecture must facilitate the learning of \textit{strong} and \textit{general} representations. 
We define a \textit{strong} representation as one that captures the discourse-level information within the entire dialog history as well as utterance-level information in the utterances that constitute that history. By our definition, a representation is sufficiently \textit{general} when it allows the model to perform better on a variety of downstream tasks. 
The next section describes the pretraining objectives within the context of the strength and generality of the learned representations. 

For clarity of discussion, the following notation is used: an arbitrary $T$-turn dialog segment is represented by a list of utterances $c=[u_1, ... u_T]$, where $u_i$ is an utterance. Further, we denote the set of all observed dialog responses in the data by $R=\{r_1, ... r_M\}$. 

The pretraining objectives, discussed below, are next-utterance retrieval (NUR), next-utterance generation (NUG), masked-utterance retrieval (MUR), and inconsistency identification (InI).

\subsection{Next-Utterance Retrieval}

NUR has been extensively explored both as an independent task \citep{lowe2015ubuntu,lowe2016evaluation} and as an auxiliary loss in a multi-tasking setup \citep{wolf2019transfertransfo}. Given a dialog context, the aim of NUR is to select the correct next utterance from a set of $k$ candidate responses. NUR can be thought of as being analogous to language modelling, except that the utterances, rather than the words, are the indivisible atomic units. Language modelling pretraining has produced strong representations of language \citep{radford2018improving, peters2018deep}, thereby motivating the choice of NUR as a pretraining objective.  

For this task we use a hierarchical encoder to produce a representation of the dialog context by first running each utterance independently through a Bidirectional Long-short Term Memory Network (biLSTM) and then using the resulting utterance representations to produce a representation of the entire dialog context. We use a single biLSTM to encode candidate responses. Given $[u_1, ... u_{T-1}]$, the task of NUR is to select the correct next utterance $u_T$ from $R$. Note that for large dialog corpora, $R$ is usually very large and it is more computationally feasible to sample a subset of $R$ and as such we retrieve $K$ negative samples for each training example, according to some distribution $p_n(r)$, e.g. uniform distribution~\cite{mikolov2013distributed}. Concretely, we minimize the cross entropy loss of the next utterance by:
\begin{align}
    \mathbf{\hat{u}_i} &= f_{\text{u}}(u_i) \quad i \in [1, T-1]\\
    [\mathbf{h_1}, ...\mathbf{h_{T-1}}]  &= f_{\text{c}}(\mathbf{\hat{u}_1},...\mathbf{\hat{u}_{T-1}})\\ 
    \mathbf{r_{gt}} &= f_{\text{r}}(u_{T})\\  
    \mathbf{r_{j}} &= f_{\text{r}}(r_{j}) \quad r_j \sim p_n(r)\\ 
    \alpha_{gt} &= \mathbf{h_{T-1}}^T \mathbf{r_{gt}}\\
    \alpha_{j} &= \mathbf{h_{T-1}}^T \mathbf{r_{j}}
\end{align}
where $f_{\text{u}}$, $f_{\text{c}}$ and $f_{\text{r}}$ are three distinct biLSTM models that are to be trained. The final loss function is:
\begin{align}
    \mathcal{L} &= -\log p(u_T|u_1,...u_{T-1}) \\ \nonumber
            &= -\log \left( \frac{\exp(\alpha_{gt})}{\exp(\alpha_{gt})+\sum_{j=1}^K \exp(\alpha_{j})} \right)
    \label{eq:nextuttloss}
\end{align}



\subsection{Next-Utterance Generation}

NUG is the task of generating the next utterance conditioned on the past dialog context. Sequence-to-sequence models \citep{sutskever2014sequence,bahdanau2014neural} have been used for pretraining \citep{dai2015semi,mccann2017learned}, and have been shown to learn representations that are useful for downstream tasks \citep{adi2016fine,belinkov2017neural}.

The hierarchical recurrent encoder-decoder architecture \citep{serban2016building} was used during NUG pretraining. Although the decoder is used in pretraining, only the hierarchical context encoder is transferred to the downstream tasks. Similarly to NUR, the optimization goal of NUG is to maximize the log-likelihood of the next utterance given the previous utterances. However, it differs in that it factors the conditional distribution to word-level in an auto-regressive manner. Specifically, let the word tokens in $u_T$ be $[w_1, ... w_N]$. The dialog context is encoded as in Eq~\ref{eq:nextuttloss} with an utterance and a context biLSTM. Then the loss function to be minimized is shown in Eq~\ref{eq:nextgenloss}:
\begin{align}
    \mathcal{L} &= -\log p(u_T|u_1,...u_{T-1}) \\ 
    &= - \sum_k^N \log p(w_k|w_{<k}, \mathbf{h_{T-1}}) 
    \label{eq:nextgenloss}
\end{align}





\subsection{Masked-Utterance Retrieval}
MUR is similar to NUR: the input contains a dialog context and a set of $K$ candidate responses. The objective is to select the correct response. The difference between the two is twofold. First, one of the utterances in the dialog context has been replaced by a randomly chosen utterance. Secondly, rather than use the final context representation to select the response that should immediately follow, the goal here is to use the representation of the \textit{replacement utterance} to retrieve the correct utterance. The replacement index $t$ is randomly sampled from the dialog segment:
\begin{equation}
    t \sim \text{Uniform}[1, T]
\end{equation}
Then $u_t$ is randomly replaced by a replacement utterance $q$ that is sampled from the negative distribution $p_n(r)$ defined in NUR. Finally, the goal is to minimize the negative log-likelihood of the original $u_t$ given the context hidden state at time-stamp $t$, i.e. $-\log p(u_{gt}|u_1,...q, ... u_T)$, where $u_{gt}$ is the original utterance at index $t$.

\begin{align}
    \mathbf{\hat{u}_i} &= f_{\text{u}}(u_i) \quad i \in [1, T]\\ 
    [\mathbf{h_1}, ...\mathbf{h_{T}}]  &= f_{\text{c}}(\mathbf{\hat{u}_1},...\mathbf{\hat{u}_{T}})\\  
    \mathbf{r_{gt}} &= f_{\text{r}}(u_{gt})\\ 
    \mathbf{r_{j}} &= f_{\text{r}}(r_{j}) \quad r_j \sim p_n(r)\\
    \alpha_{gt} &= \mathbf{h_{t}}^T \mathbf{r_{gt}}\\
    \alpha_{j} &= \mathbf{h_{t}}^T \mathbf{r_{j}}
\end{align}
The final loss function is:
\begin{align}
    \mathcal{L} &= -\log p(u_t|u_1,...q,...u_{T}) \\ \nonumber
            &= -\log \left( \frac{\exp(\alpha_{gt})}{\exp(\alpha_{gt})+\sum_{j=1}^K \exp(\alpha_{j})} \right)
\label{maskuttloss}
\end{align}

MUR is analogous to the MLM objective of \citet{devlin2018bert}, which forces model to keep a distributional contextual representation of each input token. By masking entire utterances, instead of input tokens, MUR learns to produce strong representations of each utterance.

\subsection{Inconsistency Identification}

InI is the task of finding inconsistent utterances within a dialog history. Given a dialog context with one utterance replaced randomly, just like MUR, InI finds the inconsistent utterance. The replacement procedure is the same as the one described for MUR, where a uniform random index $t$ is selected in the dialog context and $u_t$ is replaced by a negative sample $q$. 

While MUR strives to create a model that finds the original utterance, given the replacement index $t$, InI aims to train a model that can identify the replacement position $t$. Specifically, this is done via:
\begin{align}
    \mathbf{\hat{u}_i} &= f_{\text{u}}(u_i) \quad i \in [1, T]\\ 
    [\mathbf{h_1}, ...\mathbf{h_{T}}]  &= f_{\text{c}}(\mathbf{\hat{u}_1},...\mathbf{\hat{u}_{T}})\\  
    \alpha_{i} &= \mathbf{h_{T}}^T \mathbf{h_i} \quad i \in [1, T]
\end{align}
Finally, the loss function is to minimize the cross entropy of the replaced index:
\begin{align}
    \mathcal{L} &= -\log p(t|u_1,...q,...u_{T}) \\ \nonumber
            &= -\log \left( \frac{\exp(\alpha_{t})}{\sum_{j=1}^T \exp(\alpha_{i})} \right)
\label{inconsistenteq}
\end{align}




This pretraining objective aims to explicitly model the coherence of the dialog, which encourages both local representations of each individual utterance and a global representation of the dialog context. We believe that this will improve the generality of the pretrained representations.




\section{Downstream Tasks}
This section describes the downstream tasks chosen to test the strength and generality of the representations produced by the various pretraining objectives. The downstream evaluation is carried out on a lexicalized version of the MultiWoz dataset \cite{budzianowski2018multiwoz}. MultiWoz contains multi-domain conversations between a Wizard-of-Oz and a human. There are 8422 dialogs for training, 1000 for validation and 1000 for testing.

\subsection{Belief State Prediction}

Given a dialog context, the task is to predict a 1784-dimensional belief state vector. Belief state prediction (BSP) is a multi-class classification task, highly dependant on strong dialog context representations. The belief state vector represents the values of 27 entities, all of which can be inferred from the dialog context. 
To obtain the 1784-dimensional label, the entity values are encoded as a one-hot encoded vector and concatenated. The entities are shown in Appendix \ref{app:bsvalues}. Performance is measured using the F-1 score for entities with non-empty values. This approach is analogous to the one used in the evaluation of Dialog State Tracking Challenge 2 \citep{henderson2014dstc2}.  

This task measures the ability of a system to maintain a complete and accurate state representation of the dialog context. With a 1784-dimensional output, the hidden representation for this task must be sufficiently general. Therefore, any pretrained representations that lack generality will struggle on belief state prediction.

\subsection{Dialog Act Prediction}

Dialog act prediction (DAP), much like belief state prediction, is a multi-label task aimed at producing a 32-dimensional dialog act vector for the system utterances. The set of dialog acts for a system utterance describes the actions that may be taken by the system. This might include: informing the user about an attraction, requesting information about a hotel query, or informing them about specific trains. There are often multiple actions taken in a single utterance, and thus this is a multi-label task. To evaluate performance on dialog act prediction, we use the F-1 score.

\subsection{Next-Utterance Generation}

NUG is the task of producing the next utterance conditioned on the dialog history. We evaluate the ability of our models to generate system utterances using BLEU-4 \citep{papineni2002bleu}. This task requires both a strong global context representation to initialize the decoder's hidden state and strong local utterance representations.

\subsection{Next-Utterance Retrieval}

Given a dialog context, NUR selects the correct next utterance from a set of $k$ candidate responses. Though this task was not originally part of the MultiWoz dataset, we construct the necessary data for this task by randomly sampling negative examples. This task is underlined by \citet{lowe2016evaluation}'s suggestion that using NUR for evaluation is extremely indicative of performance and is one of the best forms of evaluation. Hits@1 (H@1) is used to evaluate our retrieval models. The latter is equivalent to accuracy.

Although some of these pretraining models had a response encoder, which would have been useful to transfer to this task, to ensure a fair comparison of all of the methods, we only transfer the weights of the context encoder.

\section{Experiments and Results}
This section presents the experiments and results aimed at capturing the capabilities and properties of the above pretraining objectives by evaluating on a variety of downstream tasks. All unsupervised pretraining objectives are trained on the full MultiWoz dataset \citep{budzianowski2018multiwoz}. Data usage for downstream fine-tuning differs, depending on the property being measured.

\subsection{Experimental Setup}


Each model was trained for 15 epochs, with the validation performance computed at each epoch. The model achieving the highest validation set performance was used for the results on the test data. The hyperparameters and experimental settings are shown in the Appendix \ref{sec:hyperparameters}. The source code will be open-sourced when this paper is released.

In the experiments, the performance on each downstream task was measured for each pretraining objective. Combinations where the pretraining objective is the same as the downstream task were excluded. 

The pretraining and finetuning is carried out on the same dataset. This evaluates the pretraining objectives as a means of extracting \textit{additional information} from the same data, in contrast to evaluating their ability to benefit from additional data. Though pretraining on external data may prove to be effective, identifying a suitable pretraining dataset is challenging and this approach more directly evaluates the pretraining objectives.


\subsection{Performance on Full Data}

To first examine whether the pretraining objectives facilitate improved performance on downstream tasks a baseline model was trained for each downstream task, using the entire set of MultiWoz data. The first row of Table \ref{tab:full} shows the performance of randomly initialized models for each downstream task. To evaluate the full capabilities of the pretraining objectives above, the pretrained models were used to initialize the models for the downstream tasks.

Results are shown on Table \ref{tab:full}. This experimental setup speaks to the strength and the generality of the pretrained representations. Using unsupervised pretraining, the models produce dialog representations that are strong enough to improve downstream tasks. The learned representations demonstrate generality because the multiple downstream tasks benefit from the same pretraining. Rather than learning representations that are useful for just the pretraining objective, or for a single downstream task, the learned representations are general and beneficial for multiple tasks.

\begin{table}[]
\centering
\begin{tabular}{|l|c|c|c|c|}
\hline
     & BSP            & DAP            & NUR            & NUG            \\ \cline{2-5} 
     & F-1            & F-1            & H@1            & BLEU           \\ \hline
None & \textbf{18.48} & 40.33          & 63.72          & 14.21          \\
NUR  & 17.80          & 43.25          & --             & 15.39          \\
NUG  & 17.96          & 42.31          & \textbf{67.34} & --             \\
MUR  & 16.76          & \textbf{44.87} & 62.38          & 15.27          \\
InI  & 16.61          & \textbf{44.84} & 62.62          & \textbf{15.52} \\ \hline
\end{tabular}
\caption{Results of evaluating the chosen \textit{pretraining objectives}, preceded by the baseline, on the four \textit{downstream tasks}. This evaluation used all of the training data for the downstream tasks as described in Section 5.2.}
\label{tab:full}
\end{table}

For the DAP and NUG downstream tasks, the pretrained models consistently outperformed the baseline. InI has the highest BLEU score for NUG. This may be a consequence of the importance of both global context representations and local utterance representations in sequence generation models. Both InI and MUR score much higher than the baseline and the other methods for DAP, which may be due to the fact that these two approaches are trained to learn a representation of each utterance rather than just an overall context representation. NUR has significant gains when pretraining with NUG, possibly because the information that must be captured to generate the next utterance is similar to the information needed to retrieve the next utterance. Unlike the other downstream tasks, BSP did not benefit from pretraining. A potential justification of this result is that due to the difficulty of the task, the model needs to resort to word-level pattern matching. The generality of the pretrained representations precludes this.

\subsection{Convergence Analysis}

\begin{figure*}[h!]
    \centering
    \includegraphics[width=0.24\textwidth]{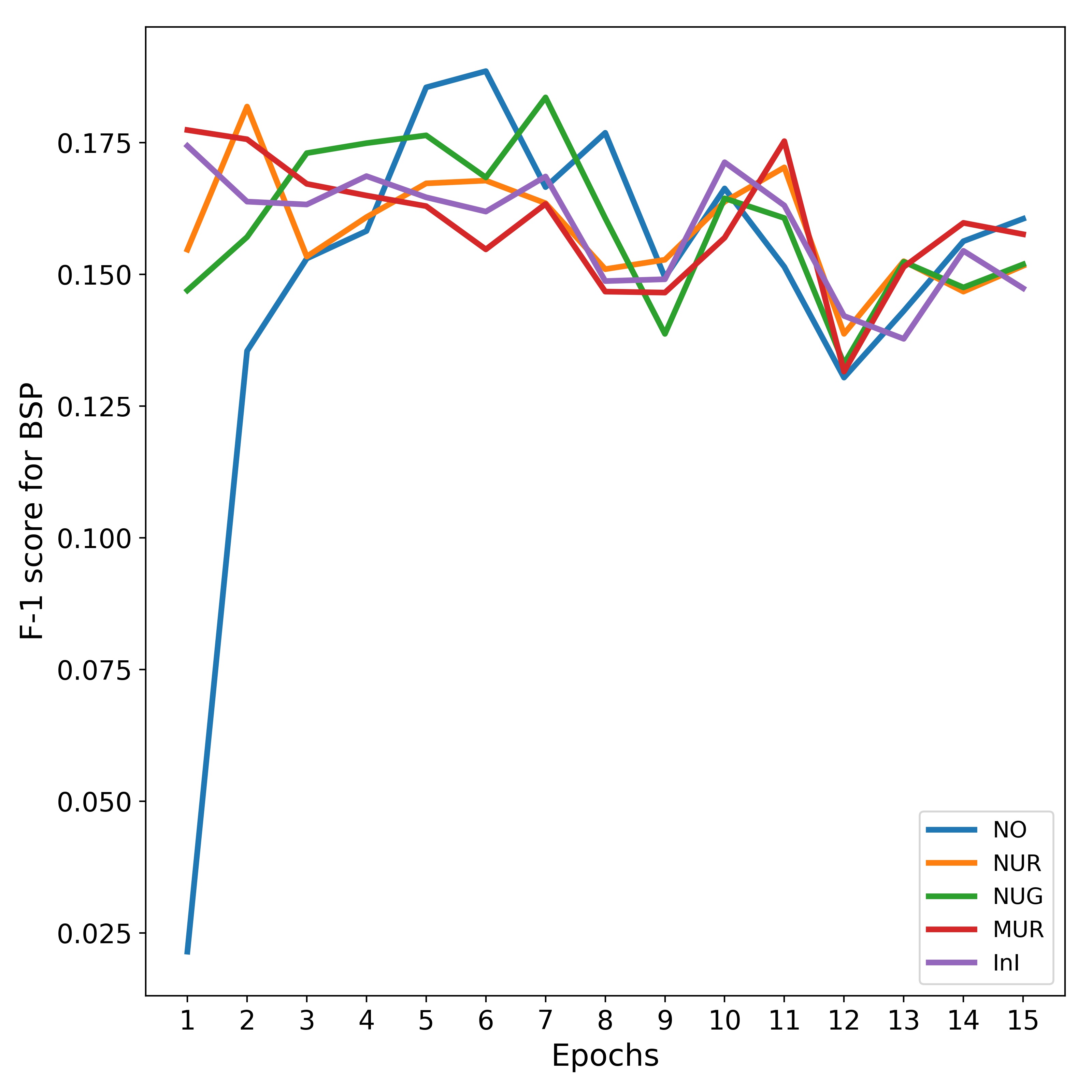}
    \includegraphics[width=0.24\textwidth]{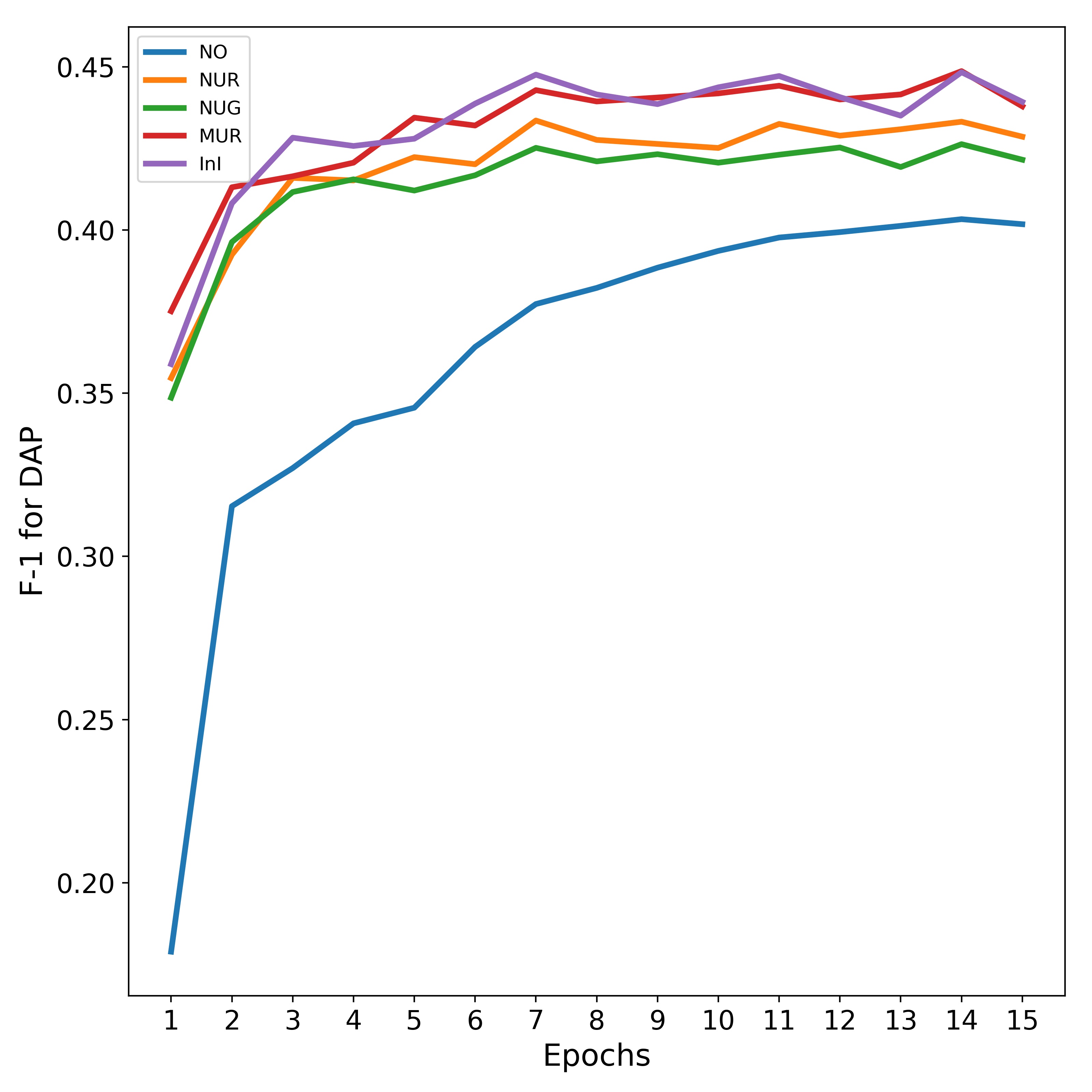}
    \includegraphics[width=0.24\textwidth]{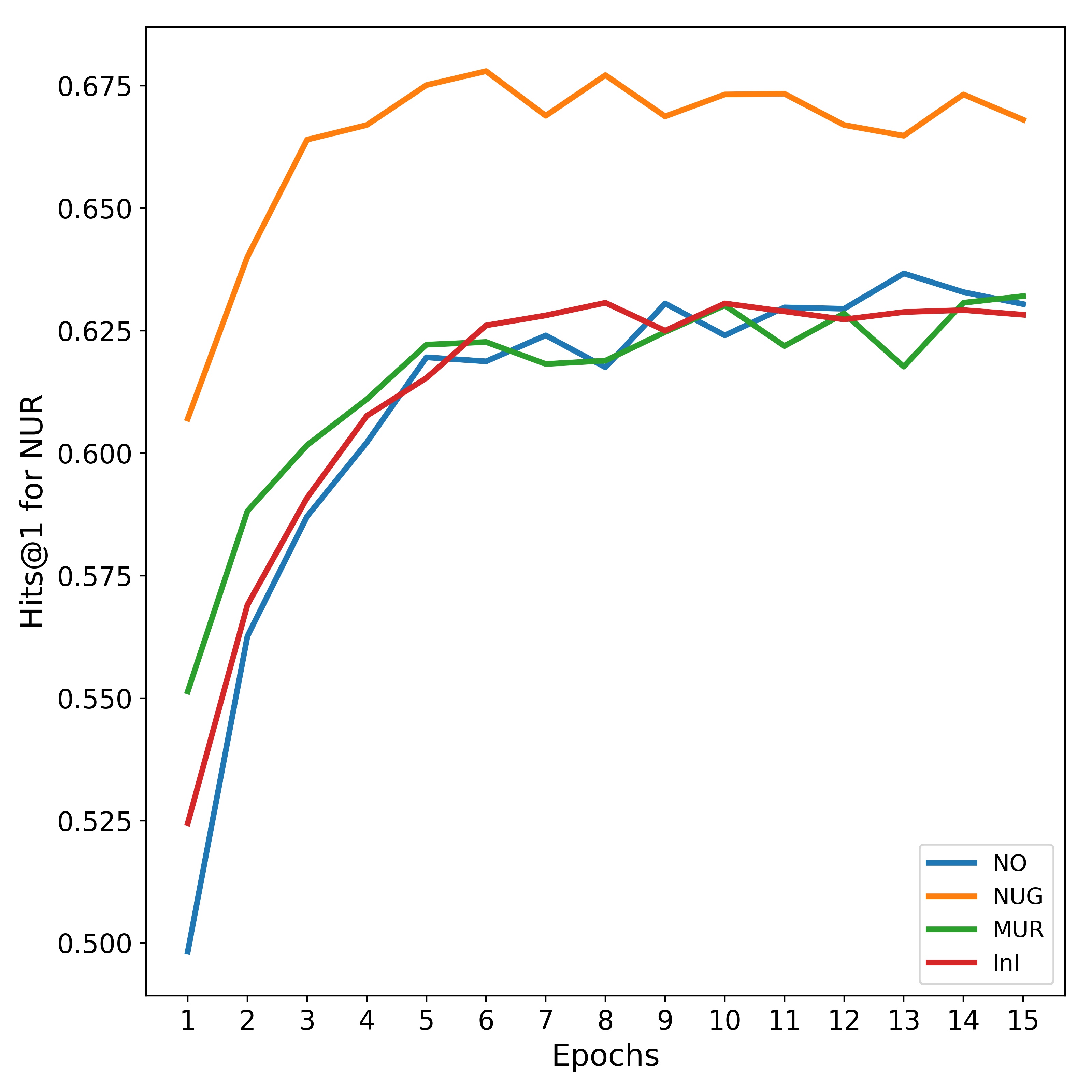}
    \includegraphics[width=0.24\textwidth]{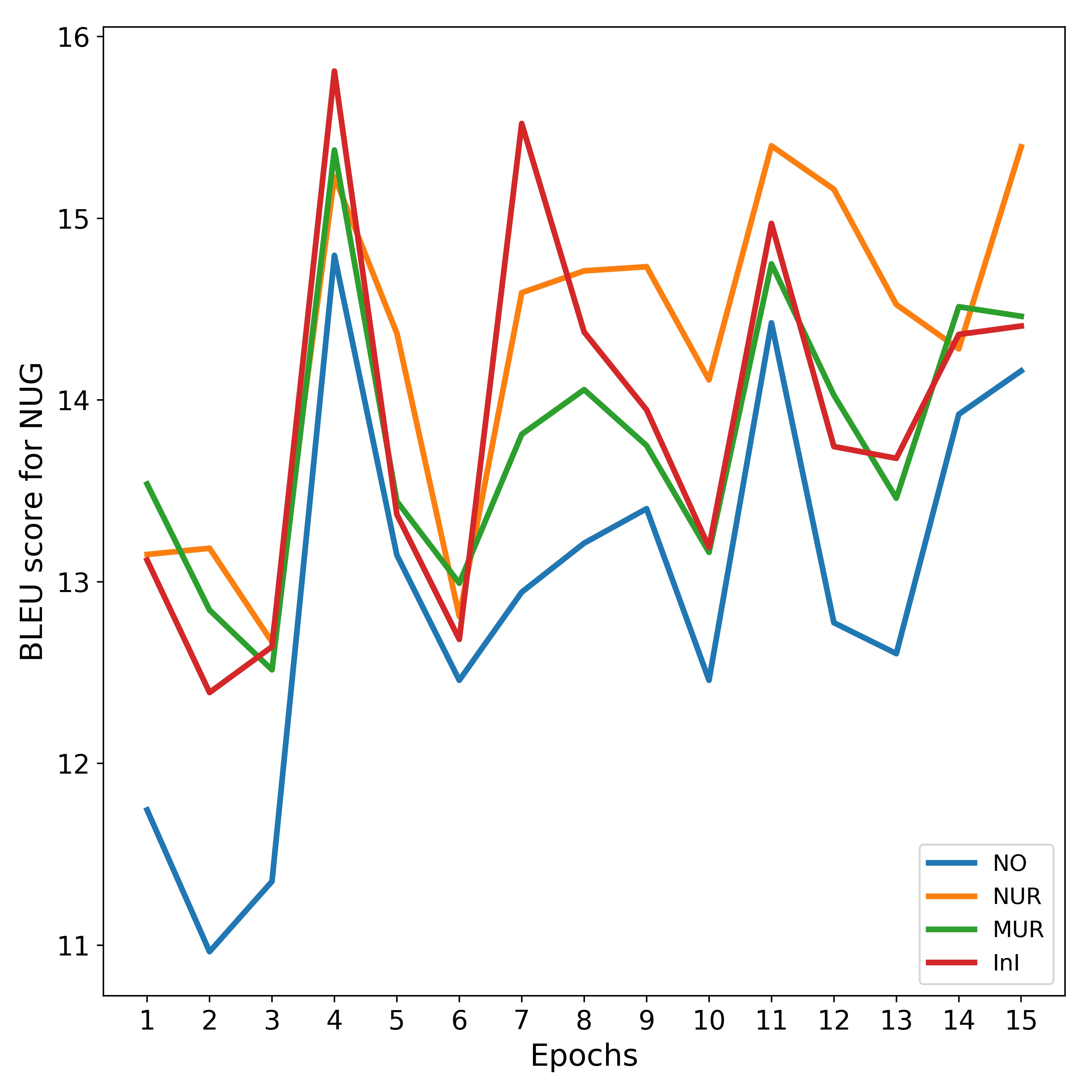}
    \caption{The performance of (from left to right) BSP, DAP, NUR, NUG across epochs with different pretraining objectives. For the BLEU-4 score in NUG, the results are noisy due to the metric being the BLEU score, however the general trend is still apparent.}
    \label{fig:bspconvergence}
\end{figure*}

This experimental setup measures the impact of pretraining on the convergence of the downstream training. Sufficiently general pretraining objectives should learn to extract useful representations of the dialog context. Thus when fine-tuning on a given downstream task, the model should be able to use the representations it has already learned rather than having to learn to extract relevant features from scratch. The performance on all downstream tasks with the different pretraining objectives is evaluated at every epoch. The results are presented on 
Figure~\ref{fig:bspconvergence}.

These figures show faster convergence across all downstream tasks with significant improvement over a random initialization baseline. The results show that performance on the initial epochs is considerably better with pretraining than without. In most cases, performance evens out during training, thus attaining results that are comparable to the pretraining methods on the full dataset. It is important to note that performance of the models after just a single epoch of training is significantly higher on all downstream tasks when the encoder has been pretrained. This underlines the usefulness of the features learned in pretraining.

The convergence of BSP shown in Figure \ref{fig:bspconvergence} is very interesting. Though the baseline ultimately outperforms all other methods, the pretrained models attain their highest performance in the early epochs. This suggests that the representations learned in pretraining are indeed useful for this task despite the fact that they do not show improvement over the baseline. 


\subsection{Performance on Limited Data}

\begin{figure}[]
    \centering
    \includegraphics[width=0.4\textwidth]{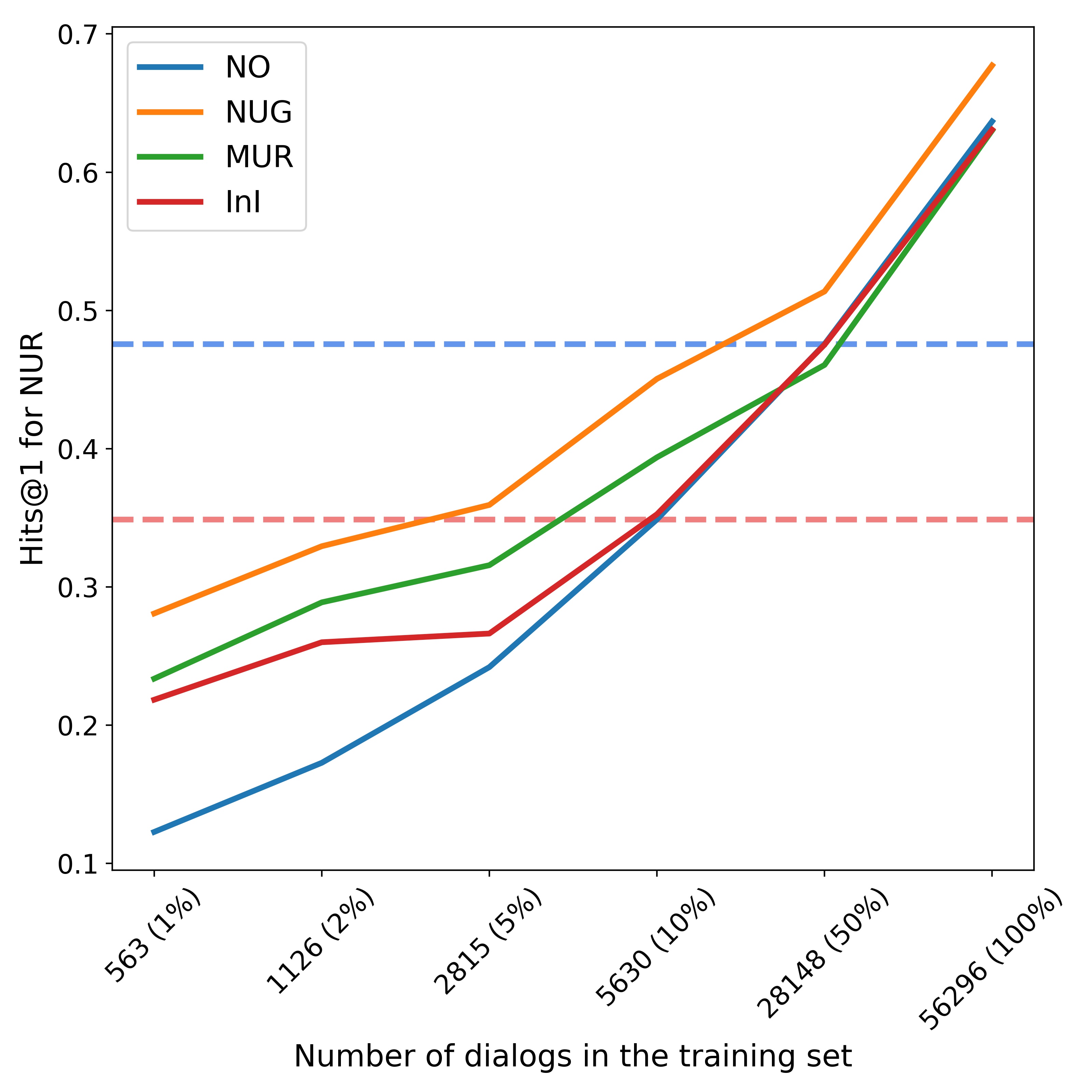}
    \caption{NUR Hits@1 at different training set sizes. The blue horizontal line is the baseline performance with 50\% of the data. The red horizontal line is the baseline performance with 10\% of the data. }
    \label{fig:less data}
    
    \vspace{-0.5em}
\end{figure}

Sufficiently strong and general pretrained representations, should continue to succeed in downstream evaluation even when fine-tuned on significantly less data. The performance on downstream tasks is evaluated with various amounts of fine-tuning data (1\%, 2\%, 5\%, 10\% and 50\%). 

The effect of the training data size for each downstream task is also evaluated. The performance of NUR with different amounts of training data is shown on Figure \ref{fig:less data}. With 5\% of the fine-tuning data, the NUG pretrained model outperforms the baseline that used 10\%. With 10\% of the fine-tuning data, this model outperforms the baseline that used 50\% of the data. 

Table \ref{tab:1results} shows all of the results with 1\% of the fine-tuning data, while Table \ref{tab:10 results} shows the results with 10\% of the fine-tuning data. More results may be found in the Appendix \ref{app: low resource}.    

\begin{table}[h!]
\centering
\begin{tabular}{|l|c|c|c|c|}
\hline
    & BSP           & DAP            & NUR            & NUG            \\ \cline{2-5} 
    & F-1           & F-1            & H@1            & BLEU           \\ \hline
None  & 4.65          & 16.07          & 12.28          & 6.82           \\
NUR & 6.44          & 14.48          & --             & \textbf{11.29} \\
NUG & \textbf{7.63} & \textbf{17.41} & \textbf{28.08} & --             \\
MUR & 5.89          & 17.19          & 23.37          & 10.47          \\
InI & 6.18          & 12.20          & 21.84          & 11.10          \\ \hline
\end{tabular}
\caption{Performance using 1\% of the data; the rows correspond to the pretraining objectives and the columns correspond to the downstream tasks. }
\label{tab:1results}
\end{table}

\begin{table}[h!]
\centering
\begin{tabular}{|l|c|c|c|c|}
\hline
    & BSP           & DAP   & NUR            & NUG            \\ \cline{2-5} 
    & F-1           & F-1   & H@1            & BLEU           \\ \hline
None  & 5.73          & 18.44 & 34.88          & 9.19           \\
NUR & 7.30          & 20.84 & --             & \textbf{14.04} \\
NUG & \textbf{9.62} & 22.11 & \textbf{45.05} & --             \\
MUR & 7.08          & \textbf{22.24} & 39.38          & 11.63          \\
InI & 7.30          & 20.73 & 35.26          & 13.23          \\ \hline
\end{tabular}
\caption{Results with 10\% of the data; the rows correspond to the pretraining objectives and the columns correspond to the downstream tasks.}
\label{tab:10 results}
\vspace{-0.5em}
\end{table}


The results shown here strongly highlight the effectiveness of pretraining. With a small fraction of the data, unsupervised pretraining shows competitive performance on downstream tasks. 

When the amount of data is very limited, the best results were obtained by models pretrained with NUG. This may be indicative of the generality of NUG pretraining. Since the generation task is difficult, it is likely that the pretrained model learns to capture the most general context representation that it can. This makes the representations especially suitable for low resource conditions since NUG pretrained representations are general enough to adapt to different tasks given even very small amounts of data,  

\subsection{Domain Generalizability}

Sufficiently general pretrained representations should facilitate domain generalizability on the downstream tasks, just as pretraining should encourage the downstream models to use domain agnostic representations and identify domain agnostic relationships in the data. 

This experimental setup is designed to mimic the scenario of adding a new domain as the downstream task. It assumes that there are large quantities of unlabeled data for unsupervised pretraining in all domains but that there is a limited set of labeled data for the downstream tasks. More specifically, for each downstream task there are $1000$ labeled out-of-domain examples (2\% of the dataset) and only $50$ labeled in-domain examples (0.1\% of the dataset). The performance of the downstream models is computed only on the in-domain test samples, thereby evaluating the ability of our models to learn the downstream task on the limited in-domain data. The results on Table \ref{tab:domain} show that pretraining produces more general representations and facilitates domain generalizability.

\begin{table}[]
\centering
\begin{tabular}{|l|c|c|c|c|}
\hline
     & BSP            & DAP            & NUR            & NUG            \\ \cline{2-5} 
     & F-1            & F-1            & H@1            & BLEU           \\ \hline
None & 4.07           & 15.22          & 13.62          & 7.80           \\
NUR  & \textbf{19.64} & 17.88          & --             & 9.97           \\
NUG  & 17.11          & \textbf{20.53} & \textbf{21.57} & --             \\
MUR  & 15.84          & 17.45          & 21.06          & 9.81           \\
InI  & 14.61          & 15.56          & 19.80          & \textbf{10.87} \\ \hline
\end{tabular}
\caption{Results of evaluating pretrained objectives on their capacity to generalize to the \textit{restaurant} domain using only 50 in-domain samples and 2000 out-of-domain samples during training. The evaluation is carried out only on the in-domain test samples.}
\label{tab:domain}
\vspace{-0.5em}
\end{table}

\section{Discussion}
The results with different experimental setups demonstrate the effectiveness of the pretraining objectives. Pretraining improves performance, leads to faster convergence, works well in low-data scenarios and facilitates domain generalizability. We now consider the respective strengths of the different pretraining objectives.

\textbf{NUR and NUG are complementary tasks.} Over all of the results, we can see that pretraining with either NUG or NUR, gives strong results when fine-tuning on the other one. This pro- perty, which has also been observed by \citet{wolf2019transfertransfo}, is a consequence of the similarity of the two tasks. Both for retrieval and generation, context encoding must contain all of the information that is necessary to produce the next utterance. 


\textbf{NUG learns representations that are very general.}  We see that NUG, especially in low data experiments, effectively transfers to many downstream tasks. This speaks to the generality of its representations. To auto-regressively generate the next utterance, the context encoder in NUG must capture a strong and expressive representation of the dialog context. This representation is all that the decoder uses to generate its response at word level so it must contain all of the relevant information. Despite the similarity of NUG and NUR, generation is a more difficult task, due to the potential output space of the model. As such, the representations learned by NUG are more general and expressive. The representative capabilities of the encoder in a generation model are also demonstrated by the work of \citet{adi2016fine}.

\textbf{InI and MUR learn strong local representations of each utterance.} The two novel pretraining objectives, InI and MUR, consistently show strong improvement for the downstream NUG task. Both of these objectives learn local representations of each utterance in the dialog context since both of their respective loss functions use the representation of each utterance instead of just the final hidden state. In an effort to better understand the properties of the different objectives, Table \ref{tab:analysis} shows performance on the NUG task for different dialog context lengths. 

\begin{table}[]
\centering
\begin{tabular}{|l|c|c|c|}
\hline
 & $< 3$    & $\geq 3$ \& $< 7$ & $\geq 7$ \\ \hline
None                  & 11.02 &         14.17    & 15.30   \\
NUR                 & \textbf{13.95}             &    15.08           & 15.88   \\
MUR                 & 12.21 &       15.36       & 16.10  \\
InI                  & 11.52  &         \textbf{15.40}      & \textbf{16.63}   \\ \hline
\end{tabular}
\caption{Results on the downstream task of NUG, with different dialog context lengths ($<3$ utterances, 3-7 utterances, and $>7$ utterances.}
\label{tab:analysis}
\vspace{-0.5em}
\end{table}

Generating a response to a longer dialog context requires a strong local representation of each individual utterance. A model that does not capture strong representations of each utterance will likely perform poorly on longer contexts. For example, for a dialog in which the user requests a restaurant recommendation, in order to generate the system utterance that recommends a restaurant, the model must consider all of the past utterances in order to effectively generate the recommendation. If the local representations of each utterance are not strong, it would be difficult to generate the system output. 

The results in Table \ref{tab:analysis} demonstrate that both InI and MUR strongly outperform other methods on long contexts, suggesting that these methods are effective for capturing strong representations of each utterance. Both MUR and InI perform poorly on shorter contexts. This further demonstrates that fine-tuned NUG models learn to rely on strong utterance representations, and therefore struggle when there are few utterances. 

\textbf{Using the same dataset for pretraining and finetuning.} The pretraining objectives demonstrate large improvements over directly training for the downstream task. No additional data is used for pretraining, which suggests that the proposed objective allow the model to extract stronger and more general context representations from the same data. The reduced data experiments show that pretraining on a larger corpora (i.e., the full data), results in strong performance on smaller task-specific datasets (i.e., the reduced data). As such, it is likely that pretraining on larger external data will result in further performance gains, however, it is challenging to identify a sufficient corpus.



\section{Conclusion and Future Work}

This paper proposes several methods of unsupervised pretraining for learning \textit{strong} and \textit{general} dialog context representations, and demonstrates their effectiveness in improving performance on  downstream tasks with limited fine-tuning data as well as out-of-domain data.   
It proposes two novel pretraining objectives:  \textit{masked-utterance retrieval} and \textit{inconsistency identification} which better capture both the utterance-level and context-level information. Evaluation of the learned representations on four downstream dialog tasks shows strong performance improvement over randomly initialized baselines.



In this paper, unsupervised pretraining has been shown to learn effective representations of dialog context, making this an important research direction for future dialog systems. These results open three future research directions. First, the models proposed here should be pretrained on larger external dialog datasets. Second, it would be interesting to test the representations learned using unsupervised pretraining on less-related downstream tasks such as sentiment analysis. Finally, the addition of word-level pretraining methods to improve the dialog context representations should be explored.

\bibliography{acl2019}

\begin{thebibliography}{33}
\expandafter\ifx\csname natexlab\endcsname\relax\def\natexlab#1{#1}\fi

\bibitem[{Adi et~al.(2016)Adi, Kermany, Belinkov, Lavi, and
  Goldberg}]{adi2016fine}
Yossi Adi, Einat Kermany, Yonatan Belinkov, Ofer Lavi, and Yoav Goldberg. 2016.
\newblock Fine-grained analysis of sentence embeddings using auxiliary
  prediction tasks.
\newblock \emph{arXiv preprint arXiv:1608.04207}.

\bibitem[{Bahdanau et~al.(2015)Bahdanau, Cho, and Bengio}]{bahdanau2014neural}
Dzmitry Bahdanau, Kyunghyun Cho, and Yoshua Bengio. 2015.
\newblock Neural machine translation by jointly learning to align and
  translate.
\newblock \emph{CoRR}, abs/1409.0473.

\bibitem[{Belinkov et~al.(2017)Belinkov, Durrani, Dalvi, Sajjad, and
  Glass}]{belinkov2017neural}
Yonatan Belinkov, Nadir Durrani, Fahim Dalvi, Hassan Sajjad, and James Glass.
  2017.
\newblock What do neural machine translation models learn about morphology?
\newblock In \emph{Proceedings of the 55th Annual Meeting of the Association
  for Computational Linguistics (Volume 1: Long Papers)}, volume~1, pages
  861--872.

\bibitem[{Budzianowski et~al.(2018)Budzianowski, Wen, Tseng, Casanueva, Ultes,
  Ramadan, and Gasic}]{budzianowski2018multiwoz}
Pawe{\l} Budzianowski, Tsung-Hsien Wen, Bo-Hsiang Tseng, I{\~n}igo Casanueva,
  Stefan Ultes, Osman Ramadan, and Milica Gasic. 2018.
\newblock Multiwoz-a large-scale multi-domain wizard-of-oz dataset for
  task-oriented dialogue modelling.
\newblock In \emph{Proceedings of the 2018 Conference on Empirical Methods in
  Natural Language Processing}, pages 5016--5026.

\bibitem[{Dai and Le(2015)}]{dai2015semi}
Andrew~M Dai and Quoc~V Le. 2015.
\newblock Semi-supervised sequence learning.
\newblock In \emph{Advances in neural information processing systems}, pages
  3079--3087.

\bibitem[{Devlin et~al.(2018)Devlin, Chang, Lee, and
  Toutanova}]{devlin2018bert}
Jacob Devlin, Ming-Wei Chang, Kenton Lee, and Kristina Toutanova. 2018.
\newblock Bert: Pre-training of deep bidirectional transformers for language
  understanding.
\newblock \emph{arXiv preprint arXiv:1810.04805}.

\bibitem[{Dhingra et~al.(2017)Dhingra, Liu, Yang, Cohen, and
  Salakhutdinov}]{dhingra2016gated}
Bhuwan Dhingra, Hanxiao Liu, Zhilin Yang, William Cohen, and Ruslan
  Salakhutdinov. 2017.
\newblock Gated-attention readers for text comprehension.
\newblock In \emph{Proceedings of the 55th Annual Meeting of the Association
  for Computational Linguistics (Volume 1: Long Papers)}, volume~1, pages
  1832--1846.

\bibitem[{Dinan et~al.(2019)Dinan, Logacheva, Malykh, Miller, Shuster, Urbanek,
  Kiela, Szlam, Serban, Lowe et~al.}]{dinan2019second}
Emily Dinan, Varvara Logacheva, Valentin Malykh, Alexander Miller, Kurt
  Shuster, Jack Urbanek, Douwe Kiela, Arthur Szlam, Iulian Serban, Ryan Lowe,
  et~al. 2019.
\newblock The second conversational intelligence challenge (convai2).
\newblock \emph{arXiv preprint arXiv:1902.00098}.

\bibitem[{Ding et~al.(2017)Ding, Yu, and Jiang}]{ding2017opinionextraction}
Ying Ding, Jianfei Yu, and Jing Jiang. 2017.
\newblock Recurrent neural networks with auxiliary labels for cross-domain
  opinion target extraction.
\newblock In \emph{Thirty-First AAAI Conference on Artificial Intelligence}.

\bibitem[{Henderson et~al.(2014)Henderson, Thomson, and
  Williams}]{henderson2014dstc2}
Matthew Henderson, Blaise Thomson, and Jason~D Williams. 2014.
\newblock The second dialog state tracking challenge.
\newblock In \emph{Proceedings of the 15th Annual Meeting of the Special
  Interest Group on Discourse and Dialogue (SIGDIAL)}, pages 263--272.

\bibitem[{Howard and Ruder(2018)}]{howard2018universal}
Jeremy Howard and Sebastian Ruder. 2018.
\newblock Universal language model fine-tuning for text classification.
\newblock In \emph{Proceedings of the 56th Annual Meeting of the Association
  for Computational Linguistics (Volume 1: Long Papers)}, volume~1, pages
  328--339.

\bibitem[{Huang et~al.(2015)Huang, Xu, and Yu}]{huang2015bidirectional}
Zhiheng Huang, Wei Xu, and Kai Yu. 2015.
\newblock Bidirectional lstm-crf models for sequence tagging.
\newblock \emph{arXiv preprint arXiv:1508.01991}.

\bibitem[{Lowe et~al.(2015)Lowe, Pow, Serban, and Pineau}]{lowe2015ubuntu}
Ryan Lowe, Nissan Pow, Iulian~V Serban, and Joelle Pineau. 2015.
\newblock The ubuntu dialogue corpus: A large dataset for research in
  unstructured multi-turn dialogue systems.
\newblock In \emph{16th Annual Meeting of the Special Interest Group on
  Discourse and Dialogue}, page 285.

\bibitem[{Lowe et~al.(2016)Lowe, Serban, Noseworthy, Charlin, and
  Pineau}]{lowe2016evaluation}
Ryan Lowe, Iulian~V Serban, Mike Noseworthy, Laurent Charlin, and Joelle
  Pineau. 2016.
\newblock On the evaluation of dialogue systems with next utterance
  classification.

\bibitem[{McCann et~al.(2017)McCann, Bradbury, Xiong, and
  Socher}]{mccann2017learned}
Bryan McCann, James Bradbury, Caiming Xiong, and Richard Socher. 2017.
\newblock Learned in translation: Contextualized word vectors.
\newblock In \emph{Advances in Neural Information Processing Systems}, pages
  6294--6305.

\bibitem[{Mikolov et~al.(2013)Mikolov, Sutskever, Chen, Corrado, and
  Dean}]{mikolov2013distributed}
Tomas Mikolov, Ilya Sutskever, Kai Chen, Greg~S Corrado, and Jeff Dean. 2013.
\newblock Distributed representations of words and phrases and their
  compositionality.
\newblock In \emph{Advances in neural information processing systems}, pages
  3111--3119.

\bibitem[{Mitkov(2014)}]{mitkov2014anaphora}
Ruslan Mitkov. 2014.
\newblock \emph{Anaphora resolution}.
\newblock Routledge.

\bibitem[{Papineni et~al.(2002)Papineni, Roukos, Ward, and
  Zhu}]{papineni2002bleu}
Kishore Papineni, Salim Roukos, Todd Ward, and Wei-Jing Zhu. 2002.
\newblock Bleu: a method for automatic evaluation of machine translation.
\newblock In \emph{Proceedings of the 40th annual meeting on association for
  computational linguistics}, pages 311--318. Association for Computational
  Linguistics.

\bibitem[{Peters et~al.(2018)Peters, Neumann, Iyyer, Gardner, Clark, Lee, and
  Zettlemoyer}]{peters2018deep}
Matthew Peters, Mark Neumann, Mohit Iyyer, Matt Gardner, Christopher Clark,
  Kenton Lee, and Luke Zettlemoyer. 2018.
\newblock Deep contextualized word representations.
\newblock In \emph{Proceedings of the 2018 Conference of the North American
  Chapter of the Association for Computational Linguistics: Human Language
  Technologies, Volume 1 (Long Papers)}, volume~1, pages 2227--2237.

\bibitem[{Plank et~al.(2016)Plank, S{\o}gaard, and
  Goldberg}]{plank2016multilingual}
Barbara Plank, Anders S{\o}gaard, and Yoav Goldberg. 2016.
\newblock Multilingual part-of-speech tagging with bidirectional long
  short-term memory models and auxiliary loss.
\newblock In \emph{Proceedings of the 54th Annual Meeting of the Association
  for Computational Linguistics (Volume 2: Short Papers)}, volume~2, pages
  412--418.

\bibitem[{Radford et~al.(2018)Radford, Narasimhan, Salimans, and
  Sutskever}]{radford2018improving}
Alec Radford, Karthik Narasimhan, Tim Salimans, and Ilya Sutskever. 2018.
\newblock Improving language understanding by generative pre-training.
\newblock \emph{URL https://s3-us-west-2. amazonaws.
  com/openai-assets/research-covers/languageunsupervised/language understanding
  paper. pdf}.

\bibitem[{Rei and Yannakoudakis(2017)}]{rei2017errordetection}
Marek Rei and Helen Yannakoudakis. 2017.
\newblock Auxiliary objectives for neural error detection models.
\newblock In \emph{Proceedings of the 12th Workshop on Innovative Use of NLP
  for Building Educational Applications}, pages 33--43.

\bibitem[{Serban et~al.(2016)Serban, Sordoni, Bengio, Courville, and
  Pineau}]{serban2016building}
Iulian~Vlad Serban, Alessandro Sordoni, Yoshua Bengio, Aaron~C Courville, and
  Joelle Pineau. 2016.
\newblock Building end-to-end dialogue systems using generative hierarchical
  neural network models.
\newblock In \emph{AAAI}, volume~16, pages 3776--3784.

\bibitem[{Sutskever et~al.(2014)Sutskever, Vinyals, and
  Le}]{sutskever2014sequence}
Ilya Sutskever, Oriol Vinyals, and Quoc~V Le. 2014.
\newblock Sequence to sequence learning with neural networks.
\newblock In \emph{Advances in neural information processing systems}, pages
  3104--3112.

\bibitem[{Thornbury and Slade(2006)}]{thornbury2006conversation}
Scott Thornbury and Diana Slade. 2006.
\newblock \emph{Conversation: From description to pedagogy}.
\newblock Cambridge University Press.

\bibitem[{Trinh et~al.(2018)Trinh, Dai, Luong, and
  Le}]{trinh2018longtermdependecies}
Trieu Trinh, Andrew Dai, Thang Luong, and Quoc Le. 2018.
\newblock Learning longer-term dependencies in rnns with auxiliary losses.
\newblock In \emph{International Conference on Machine Learning}, pages
  4972--4981.

\bibitem[{Vinyals and Le(2015)}]{vinyals2015neural}
Oriol Vinyals and Quoc Le. 2015.
\newblock A neural conversational model.
\newblock \emph{arXiv preprint arXiv:1506.05869}.

\bibitem[{Wang et~al.(2018)Wang, Singh, Michael, Hill, Levy, and
  Bowman}]{wang2018glue}
Alex Wang, Amapreet Singh, Julian Michael, Felix Hill, Omer Levy, and Samuel~R
  Bowman. 2018.
\newblock Glue: A multi-task benchmark and analysis platform for natural
  language understanding.
\newblock \emph{arXiv preprint arXiv:1804.07461}.

\bibitem[{Williams et~al.(2013)Williams, Raux, Ramachandran, and
  Black}]{williams2013dialog}
Jason Williams, Antoine Raux, Deepak Ramachandran, and Alan Black. 2013.
\newblock The dialog state tracking challenge.
\newblock In \emph{Proceedings of the SIGDIAL 2013 Conference}, pages 404--413.

\bibitem[{Wolf et~al.(2019)Wolf, Sanh, Chaumond, and
  Delangue}]{wolf2019transfertransfo}
Thomas Wolf, Victor Sanh, Julien Chaumond, and Clement Delangue. 2019.
\newblock Transfertransfo: A transfer learning approach for neural network
  based conversational agents.
\newblock \emph{arXiv preprint arXiv:1901.08149}.

\bibitem[{Yu and Jiang(2016)}]{yu2016sentimentclassification}
Jianfei Yu and Jing Jiang. 2016.
\newblock Learning sentence embeddings with auxiliary tasks for cross-domain
  sentiment classification.
\newblock In \emph{Proceedings of the 2016 Conference on Empirical Methods in
  Natural Language Processing}, pages 236--246.

\bibitem[{Zhao et~al.(2017)Zhao, Zhao, and Eskenazi}]{zhao2017learning}
Tiancheng Zhao, Ran Zhao, and Maxine Eskenazi. 2017.
\newblock Learning discourse-level diversity for neural dialog models using
  conditional variational autoencoders.
\newblock In \emph{Proceedings of the 55th Annual Meeting of the Association
  for Computational Linguistics (Volume 1: Long Papers)}, volume~1, pages
  654--664.

\bibitem[{Zhou et~al.(2016)Zhou, Dong, Wu, Zhao, Yu, Tian, Liu, and
  Yan}]{zhou2016multi}
Xiangyang Zhou, Daxiang Dong, Hua Wu, Shiqi Zhao, Dianhai Yu, Hao Tian, Xuan
  Liu, and Rui Yan. 2016.
\newblock Multi-view response selection for human-computer conversation.
\newblock In \emph{Proceedings of the 2016 Conference on Empirical Methods in
  Natural Language Processing}, pages 372--381.

\end{thebibliography}
\bibliographystyle{acl_natbib}

\clearpage
\appendix

\section{Hyperparameters}
\label{sec:hyperparameters}

\begin{table}[h]
\begin{tabular}{l|r}
Hyperparameter                                                                        & Value \\ \hline
\begin{tabular}[c]{@{}l@{}}Number of units in the \\ utterance-level RNN\end{tabular} & 150   \\
\begin{tabular}[c]{@{}l@{}}Number of units in the \\ context-level RNN\end{tabular}   & 150   \\
\begin{tabular}[c]{@{}l@{}}Number of units in the\\ decoder (if used)\end{tabular}    & 150   \\
Optimizer                                                                             & Adam  \\
Learning rate                                                                         & 0.001 \\
Gradient clipping                 & 5.0 \\
Dropout                                                                               & 0.5   \\
Batch size                                                                            & 64    \\
Number of epochs                                                                      & 15 \\
Vocabulary size                                                                            & 1000          
\end{tabular}
\caption{Training and model hyperparameters}
\label{tab:hyperparams}
\end{table}

\section{Entities in BS vector}
\label{app:bsvalues}
\begin{table}[h!]
\begin{tabular}{l|l|l}
Domain     & Entities                                                                                                                                 & \# values \\ \hline
Taxi       & \begin{tabular}[c]{@{}l@{}}'leaveAt', 'destination',\\ 'departure', 'arriveBy'\end{tabular}                                              & 839       \\ \hline
Restaurant & \begin{tabular}[c]{@{}l@{}}'time', 'day', 'people', \\ 'food', 'pricerange', 'area'\end{tabular}                                         & 269       \\ \hline
Hospital   & 'department'                                                                                                                             & 52        \\ \hline
Hotel      & \begin{tabular}[c]{@{}l@{}}'stay', 'day', 'people', \\ 'area', 'parking', \\  'pricerange', 'stars', \\  'internet', 'type'\end{tabular} & 143       \\ \hline
Attraction & 'type', 'area'                                                                                                                           & 67        \\ \hline
Train      & \begin{tabular}[c]{@{}l@{}}'people', 'ticket', \\ 'leaveAt', 'destination', \\ 'day', 'arriveBy', 'departure'\end{tabular}               & 414      
\end{tabular}
\caption{Values used in the belief state}
\label{tab:belief state values}
\end{table}

\section{Low resource setting results}
\label{app: low resource}
\begin{table}[h!]
\begin{tabular}{|l|l|l|l|l|}
\hline
 & NUR   & NUG & DAP  & BSP \\ \hline
No                  & 17.28 &     7.00        & 17.04 & 4.80  \\
NUR                 & --             &   7.55            & 17.46  & 7.10 \\
NUG                 & \textbf{32.95}  &       --       & \textbf{19.67}  & \textbf{8.37}  \\
MUR                 & 28.89  &      \textbf{11.84}         & 17.98  & 5.81  \\
InI                  & 26.00 &        10.55       & 14.31 & 6.31  \\ \hline
\end{tabular}
\caption{Results with 2\% of the data}
\label{tab:2results}
\end{table}

\begin{table}[h!]
\begin{tabular}{|l|l|l|l|l|}
\hline
 & NUR  & NUG & DAP  & BSP \\ \hline
No                  & 24.20 &      8.66         & 15.11 & 5.30  \\
NUR                 & --            &       \textbf{12.43}        & 18.31  & 7.53  \\
NUG                 & \textbf{35.93} &        --       & \textbf{19.50}  & \textbf{8.23}  \\
MUR                 & 31.57 &       10.71        & 18.82   & 5.90  \\
InI                  & 26.63 &   10.84           & 15.42  & 6.40  \\ \hline
\end{tabular}
\caption{Results with 5\% of the data}
\label{tab:5 results}
\end{table}

\begin{table}[h!]
\begin{tabular}{|l|l|l|l|l|}
\hline
 & NUR  & NUG & DAP   & BSP \\ \hline
No                  & 47.57 &      13.36         & 25.12 & 7.98   \\
NUR                 & --            &          \textbf{15.30}     & 28.81   & \textbf{12.52} \\
NUG                 & \textbf{51.37} &       --        & 29.29   & 11.28 \\
MUR                 & 46.06 &         14.56      & 29.77   & 11.59\\
InI                  & 47.52 &           14.87    & \textbf{30.19}   & 11.33 \\ \hline
\end{tabular}
\caption{Results with 50\% of the data}
\label{tab:50 results}
\end{table}

\end{document}